\documentclass[10pt,twocolumn,letterpaper]{article}

\usepackage{iccv}              

\usepackage{multirow}
\usepackage{stackengine}
\usepackage{makecell}
\usepackage{xspace}

\definecolor{MyDarkBlue}{rgb}{0,0.08,1}
\definecolor{MyAqua}{rgb}{0,0.7,0.7}
\definecolor{MyDarkGreen}{rgb}{0.02,0.6,0.02}
\definecolor{MyDarkRed}{rgb}{0.8,0.02,0.02}
\definecolor{MyDarkOrange}{rgb}{0.40,0.2,0.02}
\definecolor{MyPurple}{RGB}{111,0,255}
\definecolor{MyRed}{rgb}{1.0,0.0,0.0}
\definecolor{MyGold}{rgb}{0.75,0.6,0.12}
\definecolor{MyDarkgray}{rgb}{0.66, 0.66, 0.66}

\definecolor{iccvblue}{rgb}{0.21,0.49,0.74}
\usepackage[pagebackref,breaklinks,colorlinks,allcolors=iccvblue]{hyperref}

\def\paperID{10495} 
\def\confName{ICCV}
\def\confYear{2025}

\title{Flow to the Mode: Mode-Seeking Diffusion Autoencoders \\ for State-of-the-Art Image Tokenization 
}

\makeatletter
\renewcommand\@fnsymbol[1]{\ensuremath{#1}}
\makeatother

\author{
Kyle Sargent$^1$,~
Kyle Hsu$^1$,~
Justin Johnson$^2$,~
Li Fei-Fei$^1$,~
Jiajun Wu$^1$
\\[0.5em]
\textsuperscript{1}Stanford University,~
\textsuperscript{2}University of Michigan
}

\usepackage{dsfont}

\newcommand{\modelname}{FlowMo\xspace}

\begin{document}

\maketitle
\begin{abstract}

Since the advent of popular visual generation frameworks like VQGAN and latent diffusion models, state-of-the-art image generation systems have generally been two-stage systems that first tokenize or compress visual data into a lower-dimensional latent space before learning a generative model. Tokenizer training typically follows a standard recipe in which images are compressed and reconstructed subject to a combination of MSE, perceptual, and adversarial losses. Diffusion autoencoders have been proposed in prior work as a way to learn end-to-end perceptually-oriented image compression, but have not yet shown state-of-the-art performance on the competitive task of ImageNet-1K reconstruction. We propose \modelname, a transformer-based diffusion autoencoder that achieves a new state-of-the-art for image tokenization at multiple compression rates without using convolutions, adversarial losses, spatially-aligned two-dimensional latent codes, or distilling from other tokenizers. Our key insight is that \modelname training should be broken into a mode-matching pre-training stage and a mode-seeking post-training stage. In addition, we conduct extensive analyses and explore the training of generative models atop the \modelname tokenizer. Our code and models are available at 
\href{https://kylesargent.github.io/flowmo}{https://kylesargent.github.io/flowmo}. 
\end{abstract}   
\section{Introduction}

\begin{figure}
    \hspace{-4mm}\includegraphics[width=1.06\linewidth]{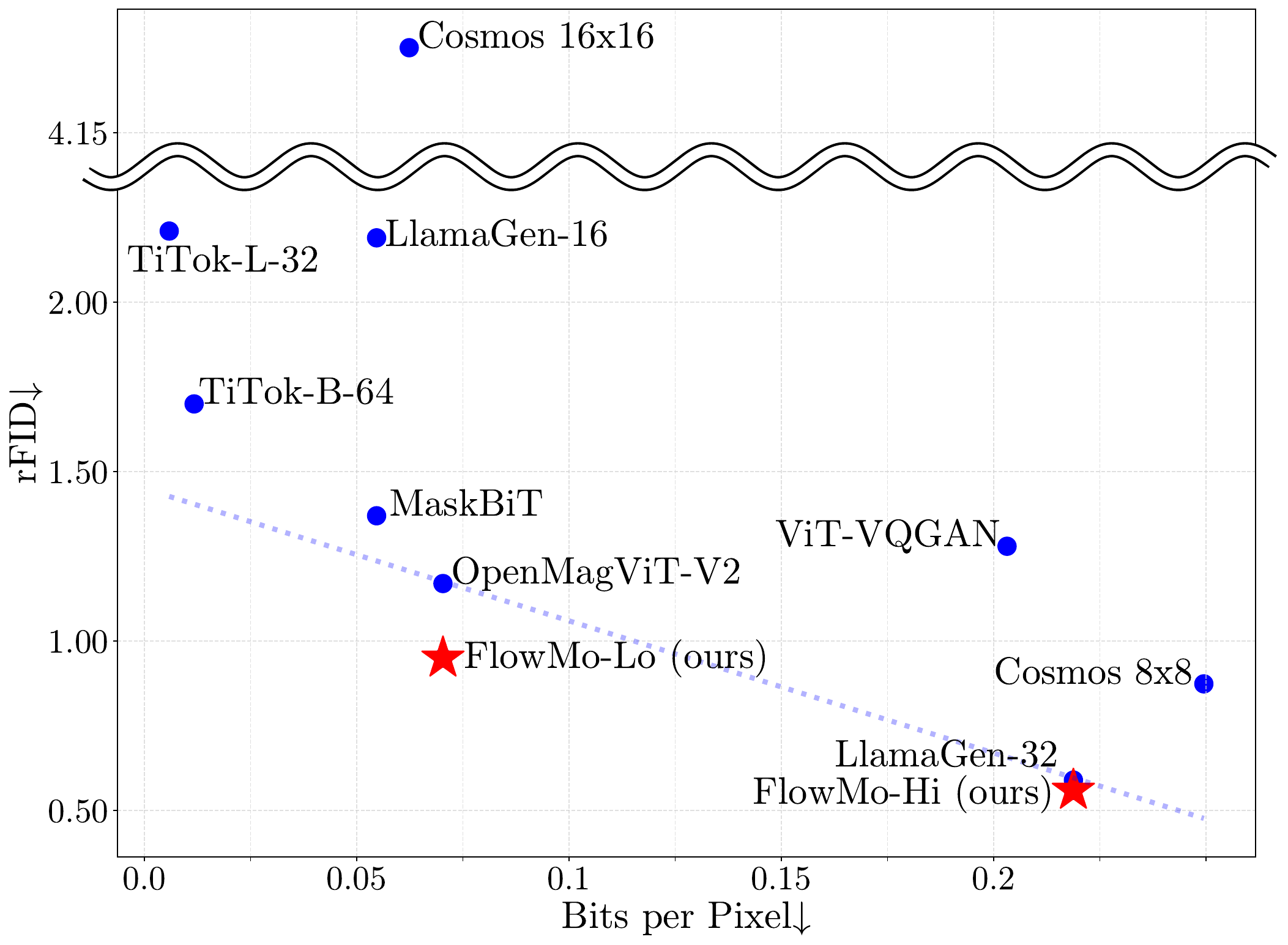}
    \vspace{-8mm}
    \caption{
        \textbf{Discrete tokenizer comparison.} State-of-the-art discrete tokenizers are benchmarked by encoding and reconstructing the ImageNet-1K validation dataset at  $256 \times 256$ resolution. with performance measured in reconstruction FID (rFID), which trades off against compression rate as measured in bits per pixel (BPP).
        Whether trained for reconstruction at a low BPP (\modelname-Lo) or high BPP (\modelname-Hi), \modelname~achieves state-of-the art image tokenization performance compared with the respective baselines. Moreover, \textit{\modelname is a transformer-based diffusion autoencoder which does not use convolutions, adversarial losses, or proxy objectives from auxiliary tokenizers.
        }
    }
    \label{fig:teaser}
    \vspace{-2mm}
\end{figure}

Generative models such as diffusion models \cite{diffusion2015,yangsonggradients,ddpm,sd3,rectified_flow,flow_matching} and discrete autoregressive models \cite{vqgan,maskgit} have seen compelling applications in the creation of image and video content \cite{sora, imagen, videpoet}. Although less common systems such as cascaded \cite{imagen} and single-stage \cite{matryoshka,simple_diffusion} systems have been explored, state-of-the-art systems for visual generation generally consist of two stages: first, a ``tokenizer'' is learned to compress pixel data to a smaller and more tractable discrete or continuous latent space; second, a generative model is trained over this compressed latent space. Due to the dominance of this two-stage paradigm, tokenization has become an active research area in its own right, with multiple influential works focusing heavily on tokenizer training and design \cite{titok, openmagvit_v2, maskbit, magvit_v2, cosmos}. In this work, we focus on tokenizers with a discrete latent space. 

\begin{figure*}[htbp]
  \centering
  \begin{tabular}{c c@{} c@{} c@{} c@{} c@{}}
    \includegraphics[width=0.19\textwidth]{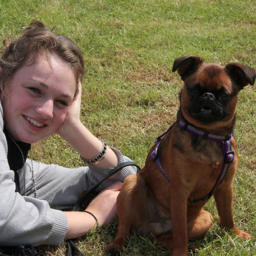} & 
    \includegraphics[width=0.19\textwidth]{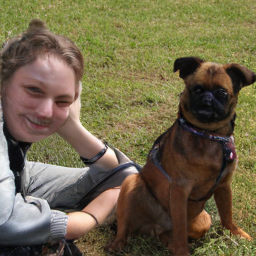} & 
    \includegraphics[width=0.19\textwidth]{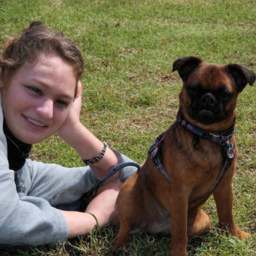} &
    \hspace{2mm}
    \includegraphics[width=0.19\textwidth]{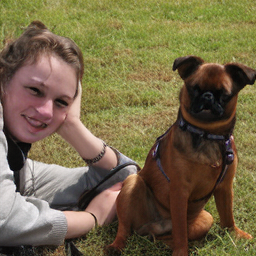} & 
    \includegraphics[width=0.19\textwidth]{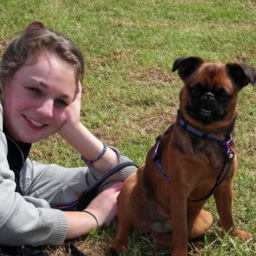} \\
    
    \includegraphics[width=0.19\textwidth]{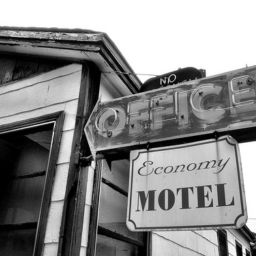} & 
    \includegraphics[width=0.19\textwidth]{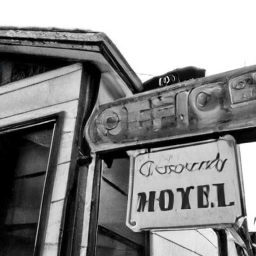} & 
    \includegraphics[width=0.19\textwidth]{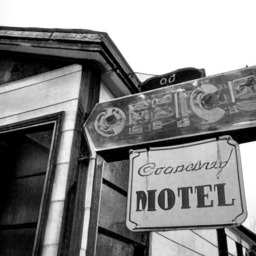} &
    \hspace{2mm}
    \includegraphics[width=0.19\textwidth]{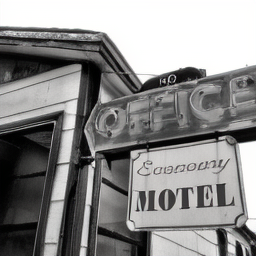} & 
    \includegraphics[width=0.19\textwidth]{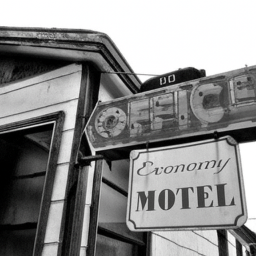} \\
    
    Original image
    
    & OpenMagViT-V2  & \modelname-Lo (Ours) & LlamaGen-32 & \modelname-Hi (Ours) \\
    & (rFID=1.17) & \textbf{(rFID=0.95)} & (rFID=0.59) & \textbf{(rFID=0.56)}
  \end{tabular}

  \vspace{-2mm}
  
  \caption{\textbf{Example reconstructions.} Comparison of original and reconstructed images of faces and text. OpenMagViT-V2 and \modelname-Lo are 0.07-bits per pixel tokenizers to be compared against each other. LlamaGen-32 and \modelname-Hi are 0.22-bits per pixel tokenizers to be compared against each other. Best viewed zoomed-in in the electronic version. More comparisons are available on \href{https://kylesargent.github.io/flowmo}{our website}.
  }
  \label{fig:comparison}
\end{figure*}

Since VQGAN \cite{vqgan}, a dominant architecture and training scheme for discrete image tokenizers has emerged. State-of-the-art image tokenizers are typically convolutional autoencoders trained to downsample visual data to a 2-dimensional, spatially-aligned latent code, and then upsample the latent code to a reconstruction regularized by adversarial and perceptual losses. 
Various deviations from this setup have been proposed: TiTok \cite{titok} used a transformer-based architecture and 1-dimensional latent code although relying on a traditional CNN-based tokenizer for an initial distillation stage, and ViT-VQGAN \cite{vitvqgan} used a transformer-based encoder and decoder. In general, however, the same basic setup for image tokenization has remained dominant.


In this work, we propose \modelname~(\textbf{Flow} to the \textbf{Mo}de), a significant technical departure from the current state-of-the-art for image tokenization. First, in tackling the task of image tokenization, we propose to model the multimodal reconstruction distribution using a rectified flow objective for the decoder \cite{rectified_flow, flow_matching}. Second, we use a fully transformer-based architecture to encode to and decode from a 1-dimensional latent code. Third, we optimize \modelname~end-to-end, without distilling from a pre-existing 2-dimensional tokenizer \cite{titok} or encoding on top of a latent space from a 2-dimensional tokenizer \cite{flextok}. 

Most importantly, \modelname~achieves state-of-the-art performance via a key insight. \modelname is a diffusion autoencoder, a system which has been explored in prior work \cite{diffae, swycc, highfidelity, yangmandt}. Yet, the state-of-the-art for the most competitive perceptual reconstruction benchmark, ImageNet-1K \cite{imagenet}, continues to be dominated by more traditional CNN- or GAN-based tokenizers. \modelname~achieves state-of-the-art tokenization via the following key insight: \textbf{for the task of perceptual reconstruction, it is better to sample modes of the reconstruction distribution that are perceptually close to the original image than to try to match all modes.} Hence, we divide \modelname~training into a \textit{mode-matching pre-training stage}, in which the system is trained end-to-end with a diffusion loss on the decoder like typical diffusion autoencoders, and a \textit{mode-seeking post-training stage}, in which the system is trained to selectively drop modes of the reconstruction distribution which are not close to the original image. We explain both stages in Section \ref{sec:method}. 

Despite its significant methodological departures from prior work, \modelname~is state-of-the-art when compared with the strongest available tokenizers at multiple BPPs. Our main contributions are as follows:


\begin{itemize}
    \item We present a simple but novel architecture for image tokenization based on diffusion autoencoders \cite{diffae} and the multimodal diffusion image transformer (MMDiT) \cite{sd3}. 

    \item We present a novel two-stage training scheme for diffusion autoencoders, consisting of \textit{mode-matching pre-training} and \textit{mode-seeking post-training}. 

    \item We set a new state-of-the-art for perceptual image tokenization in the 0.07 bits per pixel and 0.22 bits per pixel regimes, in terms of rFID, PSNR, and other metrics. We also show that a generative model trained with a \modelname tokenizer can match (though not beat) the performance of a generative model trained with a traditional tokenizer.

    \item We conduct an extensive analysis of the system design choices in \modelname, outlining several subtle but critical decisions in the noise schedule, sampler design, model design, quantization strategy, and post-training.

\end{itemize}

\vspace{-1mm}
\section{Related Work}
\vspace{-1mm}

\noindent\textbf{Image tokenization.} State-of-the-art systems for visual generation generally consist of two stages (with a few notable exceptions attempting to learn directly in pixel space \cite{simple_diffusion, simpler_diffusion, matryoshka, hourglass}). The first stage is a ``tokenizer'' which reduces the spatial or spatiotemporal dimension of pixel data by projecting to a continuous or discrete latent space. The dominance of this paradigm has led to numerous works that study tokenizers as important components in their own right \cite{dcae, titok, maskbit, magvit_v2, openmagvit_v2, llamagen}. In this work, we study tokenizers with discrete latents. Unlike prior work, \modelname~is the first diffusion autoencoder-based architecture to achieve state-of-the-art performance on ImageNet-1K reconstruction.

\noindent\textbf{Diffusion autoencoders.} Diffusion models are popular for visual generation \cite{ldm,sora}, and simplified frameworks such as rectified flow \cite{flow_matching, rectified_flow} have led to further adoption. The idea of learning an autoencoder end-to-end with a diffusion decoder was first proposed in \cite{diffae}. Various works have followed up on this idea, studying diffusion autoencoders for representation learning \cite{soda} and in particular for perceptually-oriented image compression \cite{yangmandt, swycc, dalle3}. $\epsilon-$VAE \cite{evae} explores a diffusion autoencoder objective in the context of GAN-based tokenizer trained with adversarial and perceptual losses, while DiscoDiff \cite{discodiff} uses a diffusion autoencoder to learning high-level semantic discrete latents for images prior to an autoregressive generation stage.

\noindent\textbf{Diffusion post-training.} Various work has attempted to improve the quality of diffusion model samples with dedicated post-training strategies, the goal of which is generally to instill desired attributes such as aesthetic quality. DDPO~\cite{ddpo} and DPOK~\cite{dpok} explore reinforcement learning objectives, while DRAFT~\cite{draft} and AlignProp~\cite{alignprop} explore backpropagation through the sampling chain. \modelname~uses a backpropagation-based post-training strategy, adapted to the continuous noise schedule of rectified flow and the unique setting of image tokenization.

\noindent\textbf{Concurrent work.}
In work concurrent to ours, DiTo \cite{dito} studies diffusion autoencoders for the continuous image tokenization task. Unlike DiTo, we focus on discrete image tokenization and compare against the strongest state-of-the-art discrete tokenizers available \cite{openmagvit_v2,llamagen}. Meanwhile, FlexTok \cite{flextok} proposes a system in which diffusion tokenizers are learned on top of a latent space from a traditional continuous variational autoencoder (VAE) trained with perceptual, adversarial, and reconstruction losses. \modelname does not rely on an auxiliary VAE. The reader may consult these works for a more complete picture of this developing field.
\section{Method}
\label{sec:method}

\begin{figure*}
    \centering
    \includegraphics[width=\linewidth]{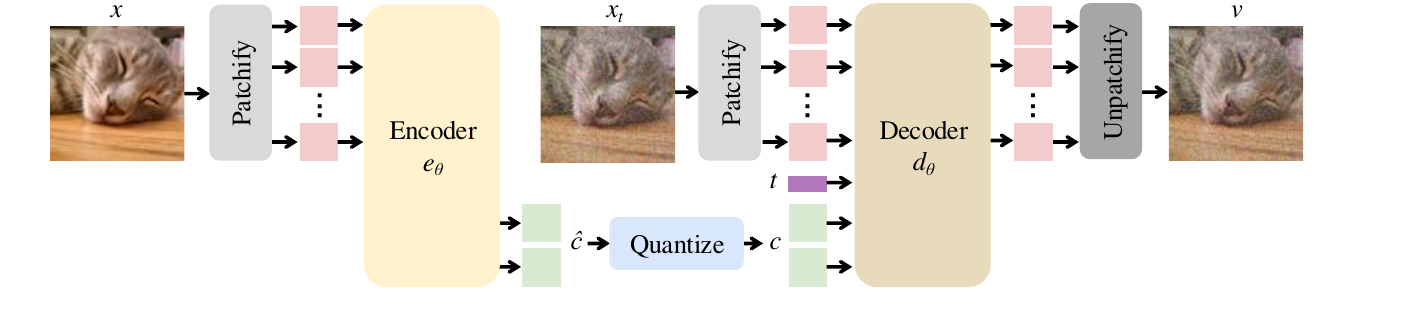}
    \vspace{-5mm}
    \caption{\textbf{\modelname architecture}. \modelname~is a diffusion autoencoder which encodes images $x$ to a latent $\hat c$ which is quantized to $c$. Then, the model decodes a rectified flow velocity $v$ conditioned on $c$ as well as a noise level $t$ and noised image $x_t$.
    }
    \label{fig:architecture}
    \vspace{-5mm}
\end{figure*}

Existing state-of-the-art tokenizers have enabled considerable progress in visual generation. However, they have a variety of drawbacks. First, they require adversarial losses \cite{magvit_v2, openmagvit_v2, vqgan}, which can be unstable and difficult to tune. They almost all require CNNs in at least one training stage, making it difficult to leverage the hardware efficiency and well-understood scaling behavior of transformers \cite{llama, mup}. Finally, they often leverage distillation from previously trained tokenizers \cite{flextok, titok} in order to achieve state-of-the-art performance.

\modelname is motivated by several key goals. We propose a tokenizer that is

\begin{itemize}
    \item \textbf{Diffusion-based.} Prior discrete tokenizers leverage adversarial losses to sample a plausible mode of the reconstruction distribution, which necessitates adaptive gradient scale computation \cite{vqgan}, LeCam regularization \cite{lecam, magvit_v2}, or careful loss weight tuning for stability. Instead, we will use a diffusion decoder, given that diffusion models have proven simple, well-suited for multi-modal modeling, and reliable at scale \cite{sora, sd3}.
    \item \textbf{Transformer-only, with 1-dimensional latent code.} Nearly every state-of-the-art image tokenization framework uses a CNN-based architecture or predicts a spatially aligned 2-dimensional latent code, with the exception of TiTok \cite{titok}, which still relies on distilling from a pre-trained CNN-based tokenizer with a 2-dimensional latent code. While these choices may provide a useful inductive bias, we feel transformer-based tokenizers with 1-dimensional latent codes may eventually provide more efficiency and flexibility at large data and model scales. \modelname learns 1-dimensional latent codes in a transformer-based architecture following MMDiT.
    \item \textbf{State-of-the-art.} There has already been considerable work in diffusion autoencoders \cite{diffae, highfidelity, swycc, soda}, but \modelname is the first to achieve state-of-the-art perceptual reconstruction on ImageNet-1K.
\end{itemize}

\modelname achieves these goals, as we will now show now, via an exposition of the system, and later (in Section \ref{sec:experiments}), via extensive experiments, ablations, and analyses. As a diffusion autoencoder, the distribution of reconstructed images $x$ given the latent
code $c$, denoted $p(x|c)$, is necessarily multimodal given the limited information in $c$. Our key insight is that in order to achieve state-of-the-art tokenization with a diffusion autoencoder, multiple steps should be taken to bias $p(x|c)$ towards modes of high perceptual similarity to the original image. We achieve this by using two key ingredients: (1) \textit{mode-seeking post-training}, explained in \ref{sec:method_posttraining}, and (2) a \textit{shifted sampler}, explained in Section \ref{sec:method_inference}. 

We will now describe \modelname in detail. First, we will go over the \modelname architecture, which is a simple transformer-based architecture based on MMDiT \cite{sd3}. Then, we will explain the two training stages: mode-matching pre-training (Stage 1A) and mode-seeking fine-tuning (Stage 1B). Finally, we will discuss generative modeling over the \modelname latent space (Stage 2) and sampling.

\subsection{Architecture}
The architectural diagram of \modelname is given in Figure \ref{fig:architecture}. 
\modelname~is a diffusion autoencoder. The encoder encodes images $x$ to a quantized latent $c$. The decoder is a conditional diffusion model which learns the conditional distribution of reconstructed images $p(x|c)$.

\modelname consists of an encoder $e_{\theta}$ and decoder $d_\theta$. Both are transformers based on the multimodal diffusion image transformer (MMDiT) \cite{sd3} adapted from Flux \cite{flux2024}. Given a patchified~\cite{vit} image $x \in \mathbb{R}^n$ and an initial latent code $c_0 \in \mathbb{R}^d$ (a vector of all zeroes), the encoder produces a latent token sequence
\begin{equation}
    \hat c = e_\theta(x, c_0),
\end{equation}
by interacting the concatenated sequence of latent and image tokens with self-attention while maintaining separate streams for each modality. $\hat{c}$ is then binarized elementwise via a quantization operation $q$ following lookup-free quantization (LFQ) \cite{magvit_v2}, yielding
\begin{equation}
    c = q(\hat c) = 2\cdot \mathds{1} [\hat c \geq 0] - 1 .
\end{equation}
Following rectified flow~\cite{rectified_flow}, the decoder is trained to model a velocity field $v$ from noise to data defined as
\begin{equation}
    v = d_\theta(x_t, c, t).
\end{equation}
The decoder processes $x_t$ and $c$ identically to how the encoder processes $x$ and $c_0$, but additionally accepts the time (or noise level) parameter $t$ to condition each MMDiT block via AdaLN modulation \cite{DiT}. The encoder $e_\theta$ and decoder $d_\theta$ are architecturally symmetric but differently sized, with the decoder being considerably larger and deeper. We use the $\mu P$ parameterization \cite{mup} for all models in order to simplify hyperparameter transfer between exploratory and scaled-up configurations.

\subsection{Stage 1A: Mode-matching pre-training}

\begin{figure} 
    \centering
    \includegraphics[width=\linewidth]{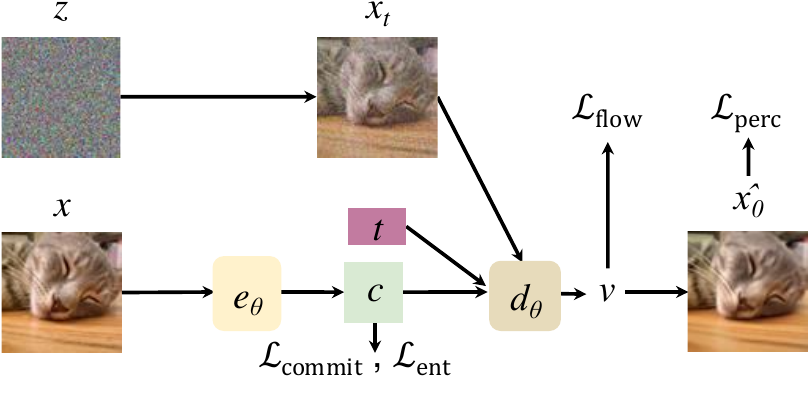}
    \caption{\textbf{Stage 1A}. The encoder and decoder are trained end-to-end with output losses $\mathcal{L}_\text{perc}, \mathcal{L}_\text{flow}$ and latent losses $\mathcal{L}_\text{commit}, \mathcal{L}_\text{ent}$.
    \label{fig:stage_1a}}
    \vspace{-3mm}
\end{figure}

In the first training stage, our goal is to train the encoder $e_\theta$ and decoder $d_\theta$ jointly so that the quantized $c$ is maximally informative about $x$, and so that $p_\theta(x|c)$ matches the true distribution $p(x|c)$, which is necessarily multimodal since the quantized $c$ contains only limited information. A full diagram of Stage 1A is given in Figure \ref{fig:stage_1a}. For theoretical justification, we appeal to prior work in diffusion autoencoders \cite{yangmandt}, which notes that the end-to-end diffusion objective corresponds to a modified variational lower bound with non-Gaussian decoder $p(x|c)$. 

\modelname is trained end-to-end as a diffusion autoencoder \cite{diffae}. Specifically, the \modelname encoder and decoder are trained end-to-end to optimize the rectified flow loss $\mathcal{L}_{\text{flow}}$ on the decoder output. Given noise $z\,{\sim}\,\mathcal{N}(0, I)$, data $x\,{\sim}\,p_x$, and time (or noise level) $t\,{\sim}\,p_t$, $t \in [0, 1]$, we define
\begin{equation}
x_t = tz + (1-t)x
\end{equation}
and we optimize the flow-matching objective
\begin{equation}
\mathcal{L}_{\text{flow}} = \mathbb{E}\bigg[
\big\|x - z - d_\theta(x_t, q(e_\theta(x)), t) \big\|_2^2 \bigg].
\end{equation}
We also use a learned perceptual distance $d_{\text{perc}}$ \cite{perceptual_loss} to supervise the 1-step denoised prediction of the network via
\begin{equation}
\mathcal{L}_{\text{perc}} = \mathbb{E}\bigg[d_{\text{perc}}(x, x_t + td_\theta(x_t, q(e_\theta(x)), t)) \bigg].
\end{equation}
Finally, on the latent code $c$ we use the entropy and commitment losses of LFQ following \cite{magvit_v2}, with
\begin{equation}
\mathcal{L}_{\text{ent}} = \mathbb{E}\big[H(q(\hat c) \big] - H(\mathbb{E}\big[q(\hat c)\big] ), 
\end{equation}
\begin{equation}
\mathcal{L}_{\text{commit}} = \mathbb{E} \bigg[ \big\|\hat c - q(\hat c) \big\|_2^2 \bigg].
\end{equation}
For simplicity, these two losses could be replaced with finite scalar quantization (FSQ) \cite{fsq}, but we find LFQ to perform slightly better. Our training loss in Stage 1A is 
\begin{equation}
    \mathcal{L}_{\text{flow}} + \lambda_{\text{perc}}\mathcal{L}_{\text{perc}} + \lambda_{\text{commit}}\mathcal{L}_{\text{commit}} + \lambda_{\text{ent}}\mathcal{L}_{\text{ent}},
\end{equation}
with loss weights and further details given in the supplementary material. Images are rescaled to lie in $[-1, 1]$. The noise level $t$ is sampled from a thick-tailed logit-normal distribution following Stable Diffusion 3 \cite{sd3}. 


\subsection{Stage 1B: Mode-seeking post-training}

\label{sec:method_posttraining}
In this stage, our goal is to optimize the decoder distribution $p_\theta(x|c)$ to seek modes which are perceptually of high similarity to the original image. We accomplish this by freezing the encoder and co-training the decoder on $\mathcal{L}_{\text{flow}}$ along with a post-training objective $\mathcal{L}_{\text{sample}}$ inspired by prior work on diffusion model post-training~\cite{draft, alignprop}.

\begin{figure}  
    \centering
    \includegraphics[width=\linewidth]{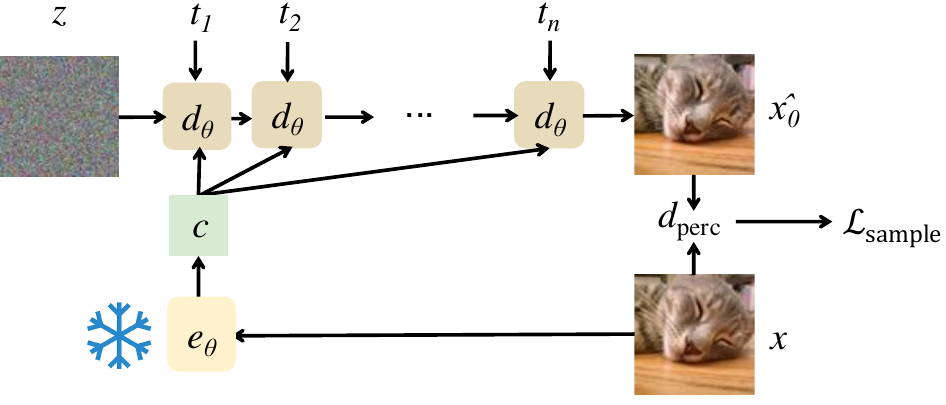}
    \caption{\textbf{Stage 1B}. The frozen encoder $e_\theta$ encodes the input image to $c$ to condition the decoder $d_\theta$, which is trained via backpropagation through the entire sampling chain. We also co-train with $\mathcal{L}_{\mathrm{flow}}$, which is the same as in Figure \ref{fig:stage_1a}. 
    \label{fig:stage_1b}}
    \vspace{-4mm}
\end{figure}

$\mathcal{L}_{\text{sample}}$ is simply a perceptual loss on the $n$-step sample from integrating the probability flow ODE, which we differentiate through. We sample the timesteps for integration $t_1, ..., t_n$ randomly to enable experimentation with different sampling schedules at test time, and use gradient checkpointing in order to backpropagate through the sampling chain. A full diagram of Stage 1B is given in Figure~\ref{fig:stage_1b}.

Let $d_{t_i}(x_t)$ denote the flow sample update function, i.e.,
\begin{equation}
    d_{t_i}(x_t) = x_t + (t_{i+1} - t_i)d_\theta(x_t, c, t_i).
\end{equation}
Then, we define
\begin{equation}
\mathcal{L}_{\text{sample}} = \mathbb{E} \bigg[  d_{\text{perc}}\left(x, d_{t_n} \circ d_{t_{n-1}} \circ\cdots \circ d_{t_1}(z   )\right)  \bigg],
\end{equation}
and our complete loss for Stage 1B is 
\begin{equation}
\mathcal{L}_{\text{flow}} + \lambda_{\text{sample}}\mathcal{L}_{\text{sample}}.
\end{equation}
We find the value of $\lambda_{\text{sample}}$ in this stage to be particularly important, and use $\lambda_{\text{sample}} = 0.01$. Too small a weight on $\mathcal{L}_{\text{sample}}$ results in poor rFID as the network forgets the perceptual features acquired in Stage 1A, whereas too large a weight results in either so-called ``reward-hacking'' as the decoder $d_\theta$ overfits to $d_{\text{perc}}$, or training divergence. We use an LPIPS-VGG network \cite{lpips} as the perceptual network in Stage 1A, but, following \cite{maskbit}, we use a ResNet as the perceptual network in Stage 1B. This and other design choices are explained in the supplementary material, and we justify there via ablation that simply using the ResNet network on the 1-step denoised prediction as in $\mathcal{L}_{perc}$ is ineffective; the post-training stage presented is necessary. This stage is computationally expensive as it requires backpropagation through the full sampling chain. We use $n=8$ and find it generalizes well to other $n$ at sampling time. 

\begin{table*}
    \centering
    \begin{tabular}{llrrrrrr}
    \toprule
    \textbf{BPP} &
    \textbf{Model} & \textbf{Tokens per image} & \textbf{Vocab size}  & \textbf{rFID$\downarrow$} & \textbf{PSNR$\uparrow$} & \textbf{SSIM$\uparrow$} & \textbf{LPIPS$\downarrow$} \\ \midrule
    0.006 &
    TiTok-L-32 \cite{titok}      & 32    & $2^{12}$    & 2.21 & 15.60 & 0.359 & 0.322  \\ \midrule
    0.012 &
    TiTok-B-64 \cite{titok}      & 64    & $2^{12}$   & 1.70 & 16.80 & 0.407 & 0.252 \\ \midrule
    0.023 &
    TiTok-S-128 \cite{titok}       & 128    & $2^{12}$   & 1.71 & 17.52 & 0.437 & 0.210 \\ \midrule
    \multirow{2}{*}{0.055} &
    LlamaGen-16 \cite{llamagen}      & 256   & $2^{14}$     & 2.19 & 20.67 & 0.589 & 0.132 \\
    & MaskBiT$^{\dagger}$ \cite{maskbit}       & 256   & $2^{14}$     & 1.37 & 21.5 & 0.56 & - \\ 
    
    \midrule
    0.062 &
    Cosmos DI-16x16 \cite{titok}       & 256    & $\approx 2^{16}$   & 4.40 & 19.98 & 0.536 & 0.153 \\ 
    \midrule
    
    \multirow{2}{*}{0.070} &
    OpenMagViT-V2 \cite{openmagvit_v2}   & 256   & $2^{18}$    & 1.17 & 21.63 & 0.640 & \textbf{0.111} \\ 
    
    & \modelname-Lo (ours) & 256   & $2^{18}$    
    & \textbf{0.95} & \textbf{22.07} & 
    \textbf{0.649} & 0.113
    
    \\ \midrule
    0.203 & 
    ViT-VQGAN$^{\dagger}$ \cite{vitvqgan}       & 1024  & $2^{13}$    & 1.28 & - & - & - \\ \midrule
    \multirow{2}{*}{0.219} & 
    LlamaGen-32 \cite{llamagen}     & 1024  & $2^{14}$    & 0.59 & 24.44 & 0.768 & \textbf{0.064} \\
    
    & \modelname-Hi (ours) & 1024  & $2^{14}$  & 
    \textbf{0.56} & \textbf{24.93} & 
    \textbf{0.785} & 0.073 \\

    \midrule
    0.249 &
    Cosmos DI-8x8 \cite{titok} & 1024    & $\approx 2^{16}$   & 0.87 & 24.82 & 0.763 & 0.072 
    \\ \bottomrule
    \end{tabular}
    \vspace{-2mm}
    \caption{\textbf{Tokenization results}. Horizontal lines separate tokenizers trained at different BPPs. \modelname achieves state-of-the-art performance at multiple BPPs compared to existing state-of-the-art tokenizers.
    $^{\dagger}$\hspace{-0.3mm}\textit{Results from the original paper.}.
    }
    \label{tab:main_results_stage1}
    \vspace{-3mm}
\end{table*}

\subsection{Sampling}
\label{sec:method_inference}
Given a quantized latent code $c$, the multimodal distribution of reconstructed images given $c$, denoted $p(x|c)$, is sampled by solving the probability flow ODE 
\begin{equation}
dx_t = v(x_t,t) = d_\theta(x_t,c,t),
\end{equation}
given an initial $x_1 \sim \mathcal{N}(0, I)$. As an aside, \modelname can also be used to compute sample log-likelihoods by solving the flow ODE in reverse following flow matching \cite{flow_matching}.

At inference, we integrate the rectified flow ODE with a timestep spacing given by a tunable shift hyperparameter $\rho$ which signifies the concentration of sampling timesteps towards lower noise levels. For a timestep spacing 
\begin{equation}
(t_1, ..., t_n) = \left( \left(\frac{n}{n}\right)^\rho, \left(\frac{n-1}{n}\right)^\rho, \dots, \left(\frac{1}{n}\right)^\rho \right),
\end{equation}
setting $\rho=1$ reduces to the usual linearly-spaced rectified flow ODE sampler. In the extreme, letting $\rho \rightarrow \infty$ means taking a single large step at $t=1$, which corresponds to regressing $x$ given $c$ \cite{genvs}. We use $\rho=4$, which corresponds to taking large steps closer to $t=1$ and concentrating the sample towards the mean of $p(x|c)$, while still spending significant sampling FLOPs at the low noise levels, a choice which prior work has shown to be critical for rFID \cite{highfidelity}. Our sampler significantly improves both rFID and PSNR.

\subsection{Stage 2: Latent generative modeling}
\label{sec:stage2}

As in other work training autoencoders with discrete latent spaces \cite{vqgan, llamagen, magvit_v2, openmagvit_v2}, we verify that our tokenizer can be used to train a high-quality second-stage generative model. We use MaskGiT \cite{maskgit} and our settings for this stage are largely taken from MaskGiT and TiTok \cite{maskgit, titok}. Additional details are in the supplementary material.
\section{Experiments}
\label{sec:experiments}

\subsection{Main results}
\textbf{Tokenization.} For the main task of tokenization, all tokenizers take an image input, encode it to a quantized latent, and then reconstruct the image. All tokenizers are trained on ImageNet-1K \cite{imagenet}. Reconstruction quality is measured in rFID \cite{frechet}, PSNR, SSIM, and LPIPS \cite{lpips}. We evaluate on the ImageNet-1K validation set, at 256$\times$256 resolution.

The amount of information contained in a tokenizer's quantized latent code is measured in bits per pixel (BPP). Where $S$ is the latent sequence length and $V$ is the token vocabulary size, the BPP of a tokenizer is computed as $\frac{S\log_2(V)}{256^2}$. Tokenizers trained at different BPP cannot be compared apples-to-apples since access to more bits will improve performance. Therefore, we train two models in order to match existing state-of-the-art tokenizers. \modelname-Lo is trained at 0.0703 BPP to match the BPP of OpenMagViT-V2 \cite{openmagvit_v2}. \modelname-Hi is trained at 0.219 BPP to match the BPP of LlamaGen-32 \cite{llamagen}. Both \modelname-Lo and \modelname-Hi are exactly the same \modelname architecture except for the BPP. To match the BPP of the tokenizer against which we compare, we modify the \modelname~token vocabulary size or number of tokens.

We achieve state-of-the-art results at both BPPs in terms of rFID \cite{frechet}, PSNR, and SSIM. The only exception to \modelname's strong performance is with respect to the LPIPS metric, for which \modelname~still tends to underperform. Extended visual comparisons are available in the supplementary material and on our website.

\begin{table}
    \setlength{\tabcolsep}{4pt} 
    \centering
    \small
    \begin{tabular}{llcccc}
    \toprule
    \textbf{Tokenizer} & \textbf{FID $\downarrow$} & \textbf{IS $\uparrow$}& \textbf{sFID $\downarrow$} & 
    \textbf{Prec. $\uparrow$} & \textbf{Rec. $\uparrow$} \\
    \midrule
    

    OpenMagViT-V2 & \textbf{3.73} & 241 & 10.66 & 0.80 & \textbf{0.51} \\
    \modelname-Lo (ours) & 4.30 & \textbf{274} & \textbf{10.31} & \textbf{0.86} & 0.47 \\
    
    \bottomrule
    \end{tabular}
    \vspace{-2mm}
    \caption{\textbf{Generation results}. We compare two MaskGiT transformers trained atop two tokenizers at the same BPP.
    }
    \label{tab:main_results_stage2}
    \vspace{-3mm}
\end{table}

\noindent\textbf{Generation.} For the task of generation, we train the generative model MaskGiT \cite{maskgit} over the encoded token sequences produced by different tokenizers. We then decode the generated token sequences via the respective tokenizers, and measure the image quality in terms of image generation metrics, namely FID \cite{frechet}, sFID \cite{sfid}, Inception Score \cite{inception_score}, and Precision and Recall \cite{precrec}. The generation metrics are evaluated on the ImageNet-1K benchmark. Despite achieving stronger tokenization, using \modelname~as a tokenizer for a downstream generative model improves some but not all generation metrics. There is an interesting and complicated interplay between tokenizer quality and generation quality, and we hope to improve the results in future work.

Training hyperparameters such as model size, batch size, and training length were identical in Table \ref{tab:main_results_stage2} save the tokenizer. Specifically, we train for 300 epochs atop both tokenizers with batch size 128 and learning rate $1 \times 10^{-4}$. Further hyperparameter settings are given in the supplementary material. While we lack the resources to train state-of-the-art generative models, which require e.g. $1,000-1,080$ epochs of training with batch size $1,024-2,048$ \cite{maskbit, titok}, our goal here is not to set a state of the art for generation, but to fairly compare tokenizers from the perspective of the latent space they provide for generation. 

\begin{figure*} 
  \centering
  \begin{tabular}{ll}
  \underline{Fully generated images - using OpenMagViT-V2 tokenizer (CFG=10.0)} \\[2mm]
  \includegraphics[width=\linewidth]{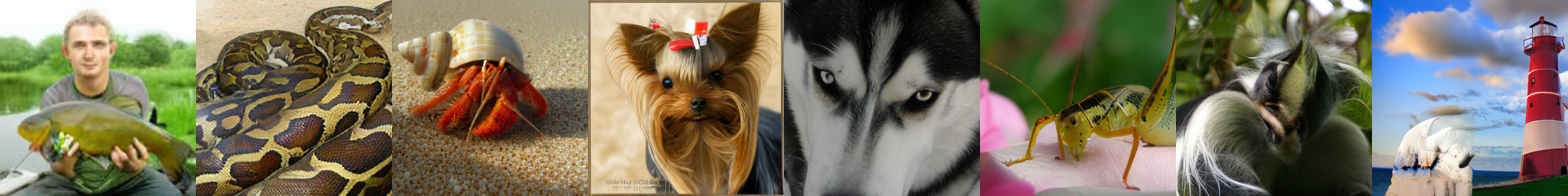} \\
  \includegraphics[width=\linewidth]{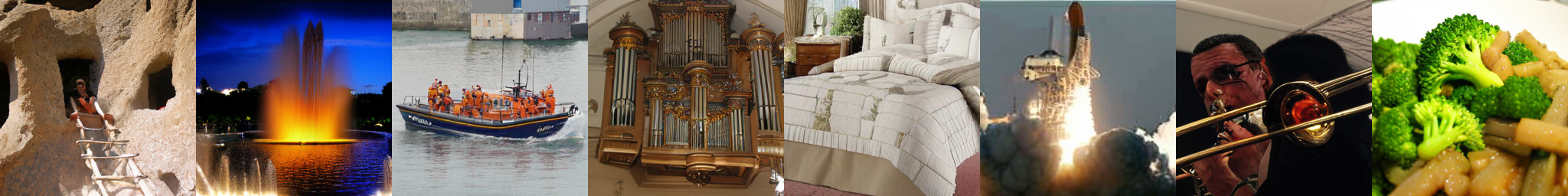} \\
  
  \underline{Fully generated images - using \modelname-Lo tokenizer (CFG=10.0)} \\[2mm]
  \includegraphics[width=\linewidth]{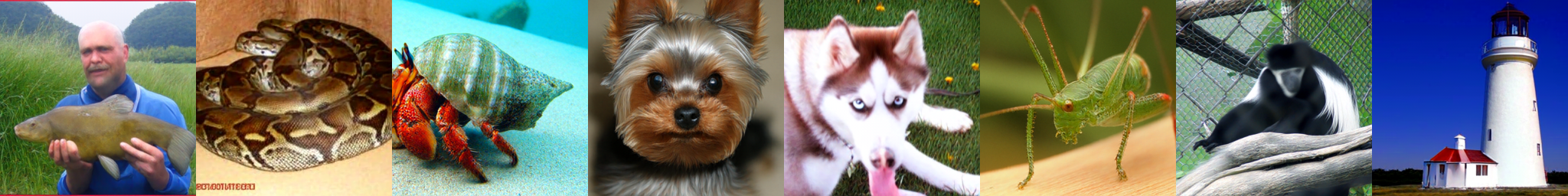} \\
  \includegraphics[width=\linewidth]{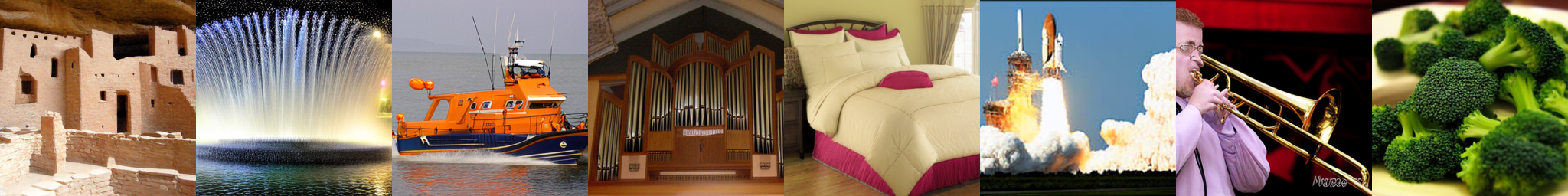} \\
  \end{tabular}
  \vspace{-3mm}
  \caption{\textbf{Generated images}. Example generated images from MaskGiT trained with different tokenizers. \modelname~can be used to train high-quality second-stage generative models. The corresponding class indices are identical for ease of comparison.}
  \vspace{-2mm}
  \label{fig:stage2_qualitative}
\end{figure*}

\begin{table}
    \setlength{\tabcolsep}{4pt} 
    \centering
    \begin{tabular}{lcccc}
    \toprule
    \textbf{Model} & \textbf{rFID $\downarrow$} & \textbf{PSNR $\uparrow$} & \textbf{SSIM $\uparrow$} & \textbf{LPIPS $\downarrow$} \\
    \midrule
    DiTo$^{\dagger}$~\cite{dito} & 0.78 & 24.10 & 0.706 & 0.102 \\
    \modelname (ours) & \textbf{0.65} & \textbf{26.61} & \textbf{0.791} & \textbf{0.054} \\
    \bottomrule
    \end{tabular}
    \vspace{-2mm}
    \caption{\textbf{DiTo comparison.} Comparison with concurrent work DiTo \cite{dito}. $^{\dagger}$\textit{Results from the original paper. 
    }}
    \label{tab:dito_comparison}
    \vspace{-4mm}
\end{table}

\noindent\textbf{Other comparisons.} We also conduct a tokenization comparison between \modelname~and DiTo \cite{dito}, a similar concurrent work, in Table \ref{tab:dito_comparison}. Exclusively for this comparison, we train a tokenizer with a continuous latent space, with LayerNorm as the final encoder layer and the ``noise sync'' augmentation as proposed in DiTo. We also equalize the overall latent size, using 256 tokens with a token dimension of 16.

\noindent\textbf{Experimental details.} All our models are parameterized in $\mu P$ \cite{mup} so that hyperparameters can be ``$\mu$-transferred'' between exploratory and scaled-up configurations. The majority of our experiments were conducted on mixed hardware (A6000, L40S, A100, H100, or H200 GPUs as available) with ViT patch size 8 and hidden dimension 768; this is also the configuration used for ablation studies in Table \ref{tab:ablation}. 

Our final experiments (\modelname-Lo, \modelname-Hi) were conducted on an 8$\times$H100 node using a scaled-up model with patch size 4 and hidden dimension increased to 1152 for the decoder only. All other hyperparameters were $\mu-$transferred directly.  We pre-train \modelname-Lo for 130 epochs and \modelname-Hi for 80 epochs, and then post-train both for approximately 1 epoch. In pre-training, performance in rFID does not saturate. Longer training would require more resources but likely further benefit performance.

We train both \modelname-Lo and \modelname-Hi with batch size 128 and learning rate $10^{-4}$. We forcibly normalize the weight matrices in the MLP blocks of the encoder and decoder per step, following EDM2 \cite{edm2}, in order to counteract exploding activations and weight matrices. All models are trained in PyTorch \cite{pytorch} with bfloat16 precision. We use the Adam optimizer \cite{adam} with $(\beta_1, \beta_2) = (0.9, 0.95)$ since we noticed higher $\beta_2$ was unstable, possibly due to bfloat16 precision or the long transformer sequence length. We set the encoder learning rate to $0$ after $200,000$ training steps. We use an exponential moving average with rate $0.9999$. 

For the full model configuration and hyperparameters, please consult the supplementary material.

\subsection{Analysis}

\noindent\textbf{Ablation studies.} We conduct ablation studies on tokenization performance, analyzing the impact of different decisions on rFID and other metrics. The setup is the same as our main ImageNet-1K tokenization experiments.


We first conduct an ablation study for design decisions in Stage 1A, in Table \ref{tab:ablation}. We analyze these decisions and reference prior diffusion autoencoder works where applicable. For these experiments, we used a configuration with reduced $\mu P$ width for efficient experimentation.

\textit{Doubled patch size.} In vision transformers \cite{vit} and derived architectures \cite{sd3}, ``patch size'' determines the length of the sequence to which an image is converted, with smaller patch sizes resulting in more computationally intensive models. Prior work on transformer-based diffusion models in pixel space has noted the importance of small patch size \cite{simpler_diffusion}. We corroborate this: increasing patch size compromises all metrics.

\textit{Encoder trained with MSE.} Prior diffusion autoencoder works have explored first training an autoencoder with an MSE- or LPIPS-based regression objective, potentially with an adversarial loss as well, and then using the resulting frozen features as conditioning for a diffusion decoder \cite{swycc,highfidelity,dalle3}. Though this style of training may boost PSNR, for best rFID it is essential to train the entire system end-to-end at all noise levels to ensure the latent code contains helpful features for velocity estimation at all noise levels.

\textit{No perceptual loss.} Prior work in diffusion autoencoders \cite{yangmandt} has noted the importance of a perceptual loss on the 1-step denoised prediction of the diffusion decoder. Without a perceptual loss, performance degrades.

\textit{FSQ quantization.} Compared with our default quantization strategy LFQ \cite{magvit_v2}, we observe that FSQ \cite{fsq} achieves slightly better final training loss and pairwise metric performance, but at the cost of rFID, which is most critical. We experimented with alternative quantization techniques as well \cite{bsq,rqvae}, but achieved best performance with LFQ.

\begin{figure}
    \centering
    \begin{tabular}{c@{}c}
    \includegraphics[width=0.48\linewidth]{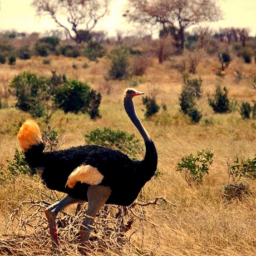}
    & \includegraphics[width=0.48\linewidth]{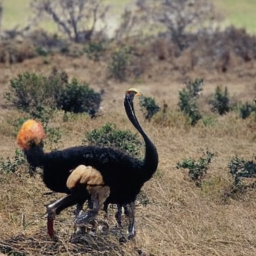} \\
    Original image & Reconstructed (discolored)
    \end{tabular}
    \vspace{-3mm}
    \caption{\textbf{Noise schedule ablation.} Training \modelname~with a logit-normal noise schedule \cite{sd3} results in discoloration.}
    \label{fig:logit-normal}
    \vspace{-6mm}
\end{figure}

\textit{Logit-normal noise.} The logit-normal noise schedule for rectified flow training \cite{sd3} assigns 0 probability mass to $t \in \{0,1\}$. 
Since our conditioning signal is extremely strong, the estimate of $v(x_t,t)$ at $t=1$ (pure noise) is critical. Using the usual logit-normal schedule causes PSNR and rFID to collapse. Instead, we use a thick-tailed logit-normal noise schedule: we sample the noise level from a uniform distribution on $[0, 1]$ $10\%$ of the time. Without this modification, the result is low PSNR and discoloration. See Figure \ref{fig:logit-normal} for an illustrative example.

\textit{Unshifted sampler.} We integrate the rectified flow ODE using a spacing of timesteps $t$, which is proportionately more concentrated towards $t=0$, as explained in Section \ref{sec:method_inference}. Using linearly-spaced timesteps degrades performance.

\textit{No guidance.} One advantage of \modelname~over a GAN tokenizer is the ability to leverage techniques such as classifier-free guidance \cite{cfg} and guidance intervals \cite{cfg_interval} to improve perceptual quality. Without applying classifier guidance in a limited interval, rFID is worsened. We enable guidance by dropping out the quantized latent  code $c$ $10\%$ of the time during tokenizer training. We do not leverage any class information in the tokenizer.

\begin{table}
\setlength{\tabcolsep}{4pt} 
    \centering
\begin{tabular}{lrcc}
\toprule
\textbf{Model} & \textbf{rFID}$\downarrow$ & \textbf{PSNR}$\uparrow$  & \textbf{LPIPS}$\downarrow$ \\ \midrule
\modelname (fewer params.)      & \textbf{2.87}  & 20.71 & 0.15        \\
\midrule
with doubled patch size & 6.39  & 19.94 & 0.17        \\
with MSE-trained encoder & 3.82 & 21.40  & 0.15       \\
without perceptual loss  & 13.86 & \textbf{22.11} & 0.21       \\
with FSQ quantization  & 3.14 & 21.31 & \textbf{0.14}       \\
with logit-normal schedule & 4.08 & 16.45 & 0.21        \\ 
without shifted sampler & 3.42 & 20.25 & 0.16  \\
without guidance & 3.28 & 20.67 & 0.16  \\
\bottomrule
\end{tabular}
\vspace{-2mm}
\caption{\label{tab:ablation} \textbf{Stage 1A Ablation}. Deviating from \modelname~design choices compromises either PSNR or rFID. We prioritize rFID in our model due to its correlation with perceptual quality.}

\vspace{-2mm}
\end{table}

\begin{figure}
\centering
    \begin{tabular}{@{}c@{}c@{}c@{}}
         \includegraphics[width=.34\linewidth]{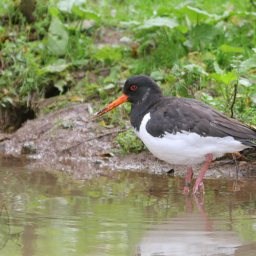} & 
         \includegraphics[width=.34\linewidth]{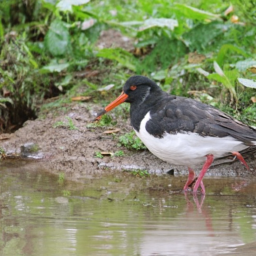} & 
         \includegraphics[width=.34\linewidth]{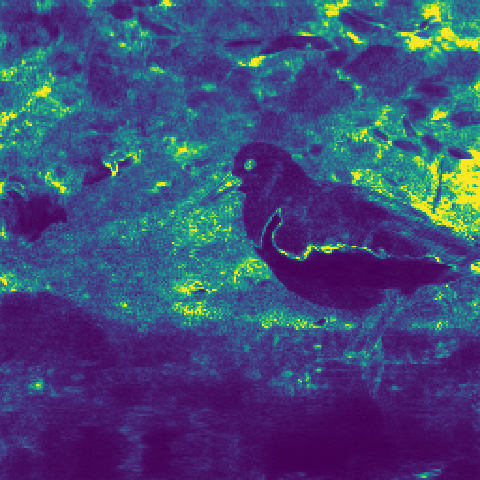} \\
         Original image & Reconstructed & Variance heatmap
    \end{tabular}
    \vspace{-3mm}
    \caption{\textbf{Multimodal reconstruction.} After post-training, \modelname~reconstruction remains multimodal, but biased towards preserving the perceptually relevant details of the image, which manifests here by the variance concentrating in the background. }
    \label{fig:posttrain_qual}
    \vspace{-5mm}
\end{figure}

Table \ref{tab:ablation_1b} is an ablation study for the inclusion of Stage 1B. Without this critical stage, performance on all metrics drops. Intriguingly, despite only training on $\mathcal{L}_{\textrm{sample}}$, this stage also improves PSNR. We illustrate in Figure \ref{fig:posttrain_qual} how the \modelname~decoder remains multimodal after post-training.

\begin{table}
    \centering
\begin{tabular}{lrccc}
\toprule
\textbf{Model} & \textbf{rFID}$\downarrow$ & \textbf{PSNR}$\uparrow$ & \textbf{LPIPS}$\downarrow$\\
\midrule

\modelname-Lo & 1.10 & 21.38 & 0.134 \\
\modelname-Lo (post-trained) & \textbf{0.95} & \textbf{22.07} & \textbf{0.113} \\
\midrule

\modelname-Hi & 0.73 & 24.02 & 0.086 \\
\modelname-Hi (post-trained) & \textbf{0.56} & \textbf{24.93} & \textbf{0.073} \\
\bottomrule
\end{tabular}
\vspace{-2mm}
\caption{\label{tab:ablation_1b} \textbf{Stage 1B ablation}. Without the post-training stage, performance is considerably worse.}
\vspace{-2mm}
\end{table}

\vspace{-4mm}
\paragraph{Limitations.}
The primary limitation of \modelname~is inference time. \modelname requires multiple model forward passes (we use $n=25$ steps in our work) to generate a sample from the decoder given a quantized latent code. 

\vspace{-1mm}
\section{Conclusion}
\vspace{-1mm}

We have presented \modelname, a transformer-based diffusion autoencoder that achieves state-of-the-art image tokenization.
We have shown state-of-the-art performance on the competitive ImageNet-1K benchmark at 256$\times$256 resolution \textit{without using 2D spatially aligned latent codes, adversarial losses, proxy objectives from auxiliary tokenizers, or convolutions}, unlike much prior work.

\noindent\textbf{Acknowledgments.} This work is in part supported by ONR MURI N00014-22-1-2740.

{
    \small
    \bibliographystyle{ieeenat_fullname}
    \bibliography{main}
}

\clearpage
\setcounter{page}{1}
\maketitlesupplementary

\appendix
\section{Training setup and hyperparameters}
\label{sec:train_setup}


\subsection{Stage 1A - Tokenizer pre-training}

All hyperparameters are given in Table \ref{tab:training_hyperparameters}. We found that for \modelname, using a relatively low commitment loss weight relative to GAN-based tokenizers was beneficial. In fact, bounding the commitment loss by applying a $\mathrm{tanh}$ activation prior to quantization degraded performance. We use token factorization following \cite{magvit_v2, openmagvit_v2}, which enables a large vocabulary size without excessive memory usage in the calculation of the entropy loss. 

Due to limited computational resources, we parameterized all \modelname~models in $\mu P$ and determined the optimal losses, noise schedules, and other hyperparameters in the exploratory ``\modelname (fewer params)'' configuration, which was trained on A100, H100, H200, L40S, or A6000 GPUs as available. We then transfer these hyperparameters to the larger and more FLOP-intensive \modelname-Lo and \modelname-Hi configurations for final experiments. We train these final configurations for relatively few epochs (130 and 80 respectively) on a single $8\times$H100 node, and it is likely that further training would continue to improve results.

\begin{table*}
    \centering
    \begin{tabular}{l|lll}
        \textbf{Hyperparameter} & \textbf{\modelname (fewer params)} & \textbf{\modelname-Lo} & \textbf{\modelname-Hi} \\
        \midrule
        Learning rate & 0.0001 & - & -  \\
        Batch size & 128 & - & - \\
        Weight decay & 0 & - & -  \\
        Num. epochs & 40 & 130 & 80 \\
        \midrule
        $\lambda_{ent}$ & 0.0025 & - & - \\
        $\lambda_{commit}$ & 0.000625 & - & - \\
        $\lambda_{lpips}$ & 0.1 & - & - \\
        \midrule
        Hidden size ($\mu P$ width) & 768 & 1152 & 1152 \\
        MLP ratio & 4 & - & - \\
        Encoder patch size & 8 & 4 & 4 \\
        Decoder patch size & 8 & 4 & 4 \\
        Encoder depth & 8 & - & - \\
        Decoder depth & 16 & - & - \\
        Latent sequence length & 256 & - & -  \\
        Latent token size & 18 & 18 & 56 \\
        Codebook size for entropy loss & 9 & 9 & 14 \\
        \midrule
        Total number of parameters $(\times 10^6)$ & 517 & 945 & 945
    \end{tabular}
    
    \caption{\label{tab:training_hyperparameters} Training hyperparameters for tokenizer training. `-' means hyperparameters are identical between \modelname~Small, A, and B configurations}
\end{table*}

\subsection{Stage 1B - Tokenizer post-training}

MaskBiT \cite{maskbit} noted the effectiveness of a ResNet-style perceptual loss for tokenizer training, and that this loss may have been used in prior work whose tokenizer training recipe was unknown \cite{maskgit}. However, we were unable to use this network effectively in our loss $\mathcal{L}_{\textrm{percep}}$, instead obtaining better results in Stage 1A with the LPIPS-VGG network. 
Interestingly, \cite{yangmandt} also applied a perceptual loss based on LPIPS-VGG on the 1-step denoised prediction of the network. 

To verify the necessity of the post-training algorithm we propose for \modelname, we consider an alternative post-training stage in which the ResNet perceptual loss is used but on the 1-step denoised prediction as in $\mathcal{L}_{\textrm{perc}}$. We sweep a range of loss weights for this ResNet-based perceptual loss in Table \ref{tab:alt_percep}, using the FlowTok-small configuration. Importantly, none are able to match the performance of $\mathcal{L}_{\textrm{sample}}$ post-training. 

\begin{table}
    \centering
    \begin{tabular}{c|ll}
        \textbf{Loss type} & \textbf{Loss weight} & \textbf{rFID} \\
        \midrule
        $\mathcal{L}_{\textrm{sample}}$ & 0.01 & 1.28 \\
        \midrule
        $\mathcal{L}_{\textrm{perc}}$ & 0.05 & 2.57 \\
        $\mathcal{L}_{\textrm{perc}}$ & 0.01 & 2.38 \\
        $\mathcal{L}_{\textrm{perc}}$ & 0.005 & 2.16\\
        $\mathcal{L}_{\textrm{perc}}$ & 0.001 & 1.67 \\
        $\mathcal{L}_{\textrm{perc}}$ & 0.0005 & 1.61 \\
        $\mathcal{L}_{\textrm{perc}}$ & 0.0001 & 1.81 \\
        \midrule
        $\mathcal{L}_{\textrm{perc}}$ & 0.00025 & 1.64 \\
        $\mathcal{L}_{\textrm{perc}}$ & 0.0005 & 1.61 \\
        $\mathcal{L}_{\textrm{perc}}$ & 0.00075 & 1.60 \\
    \end{tabular}
    \caption{$\mathcal{L}_{\textrm{sample}}$ ablation. We show the result of training with $\mathcal{L}_{\textrm{sample}}$ in the first row. We then compare with $\mathcal{L}_{\textrm{perc}}$ at various weights. We conduct two sweeps, first to find the optimal loss weight order of magnitude (middle rows) and then to fine-tune the loss weight (bottom rows). Regardless, it cannot match the performance of $\mathcal{L}_{\textrm{sample}}$ Backpropagation through a complete sampling chain is important.}
    \label{tab:alt_percep}
\end{table}

During post-training, due to the computational expense, we reduce the batch size to 64 and correspondingly reduce the learning rate to $5 \times 10^{-5}$. In addition to gradient checkpointing, we also use gradient accumulation in this stage. We train Stage 1B for approximately 1 epoch for both \modelname-Lo and \modelname-Hi, and apply early stopping to counteract eventual reward hacking \cite{ddpo, alignprop}.

We sample the timesteps to integrate the ODE for $\mathcal{L}_{\textrm{sample}}$ randomly by taking $u_1, ..., u_n \sim \textrm{Unif}(0, 1)$ and then setting 
\[
t_i = \big( \sum\limits_{j=i}^n u_j \big)/ \sum\limits_{j=1}^n u_j \big)
\]

\subsection{Stage 2 - Generative model training}

To verify the quality of an image tokenizer, it is necessary to ensure it can be used to train a high-quality generative model. As in tokenizer training, we lack resources for matching the computational requirements of state-of-the-art generative models which are trained for e.g., 1000-1080 epochs with batch size 1024-2048 \cite{titok, maskbit}. We use a MaskGiT configuration with 
hidden size 1024 and 28 layers, with 397M total parameters. training for 300 epochs with learning rate $10^{-4}$ and batch size 128. 
Both \cite{openmagvit_v2} and \modelname~produce 256 tokens of size 18 bits, which we convert to 512 tokens of size 9 bits for training MaskGiT. We use guidance 1.5 for both models, and tune the sampling temperature separately for each model to ensure fair comparison, since we find that models trained atop OpenMagViT-V2 prefer lower temperature sampling, while those trained atop \modelname~prefer high temperature. We use 64 sampling steps for both models with softmax temperature annealing and no guidance decay.

The purpose of our second stage training is to ensure \modelname~can be used to train a second stage generative model of comparable quality to one trained with a traditional tokenizer \textit{given the same computational resources.} Our goal is not to improve the state of the art for ImageNet-1K generation at 256 resolution.

\section{Evaluation setup and hyperparameters}
\label{sec:eval_setup}


For the \modelname~tokenizer, we use a guidance interval \cite{cfg_interval} of $(0.17, 1.02)$ in the EDM2 \cite{edm2} convention (translating to (0.145, 0.505) in terms of rectified flow noise levels), with classifier-free guidance weight 1.5. For the second stage, we also apply guidance, as is common practice, by shifting the token logits according to the logit difference with and without conditioning. 

We did not observe negative interaction between use of guidance in both the tokenizer and generative model; guidance in both stages is helpful.

For all evaluation metrics (rFID, PSNR, etc.) reported in the main paper, we use 50,000 samples unless otherwise noted. We compute the rFID and other reconstruction metrics in Stage 1 against the entire 50,000-example validation set. Generative metrics such as gFID are computed against the full ImageNet-1K training dataset statistics, as is standard practice. The generation results are evaluated using the ADM codebase \cite{adm}.

\section{Additional Experiments}
In this section, we provide some additional analyses and experiments to better understand and illustrate the capabilities of \modelname.

\subsection{Rate-distortion-perception tradeoff}
It could be argued that the desired ``mode-seeking'' behavior could be accomplished by simply increasing the sample likelihood. Unfortunately, evaluating the sample likelihood requires solving the log-determinant of the jacobian of the flow field \cite{flow_matching}, making it impractical to optimize directly. Neglecting the impact of this term we may increase or decrease the likelihood of $x_1$ by modulating its norm, since it is a centered isotropic gaussian (somewhat analogous to truncation sampling in GANs \cite{biggan}). 

\begin{figure*}
  \centering
  \includegraphics[width=.5\linewidth]{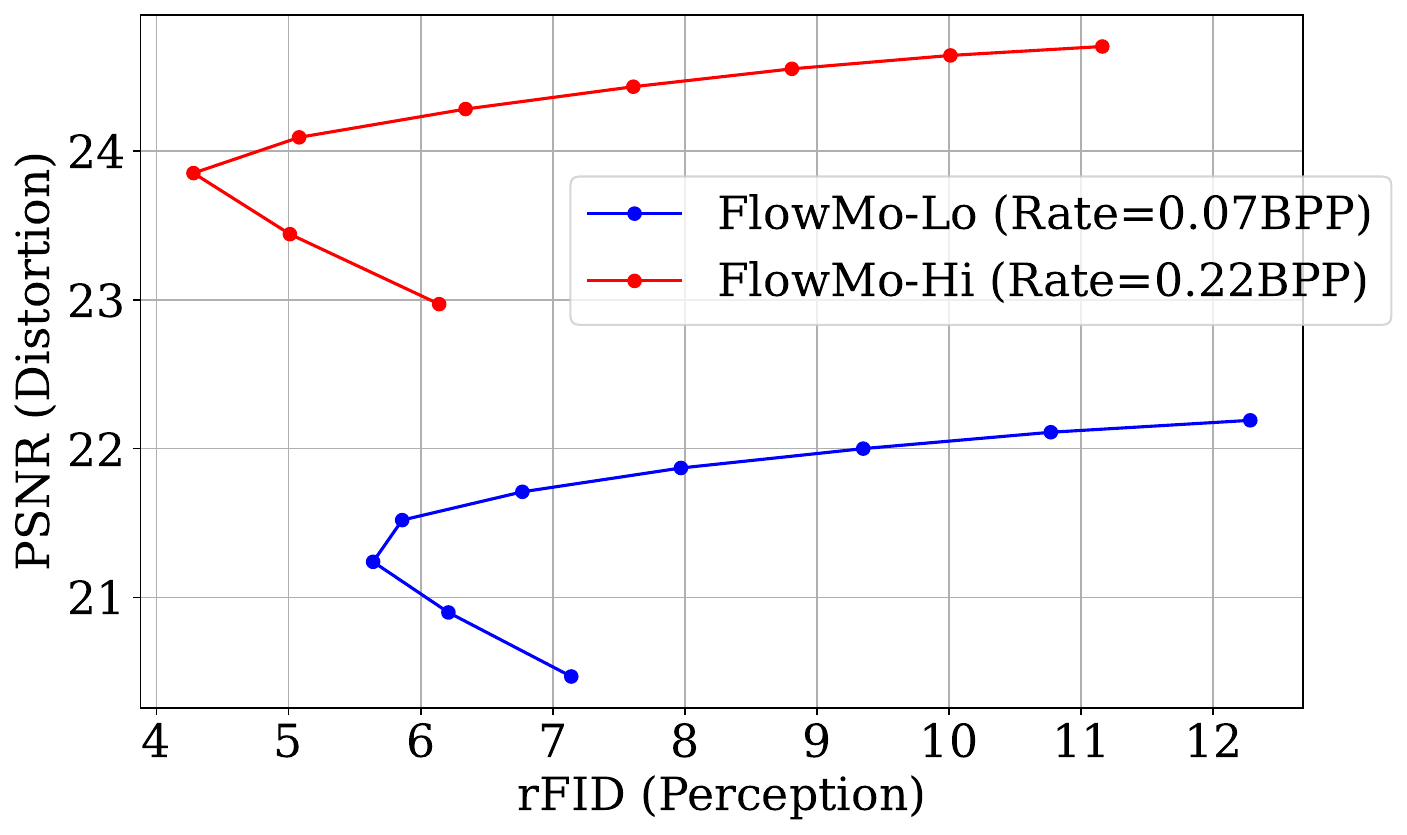}

  \caption{\label{fig:ratedistortionperception_a} Analysis of \modelname~with respect to the rate-distortion-perception tradeoff \cite{ratedistortionperception}. Varying $x_1$ likelihood sweeps out different rFID/PSNR curves at different rates. For this study, we use 5,000 samples for all metrics to make it computationally feasible.}
  
\end{figure*}

We attempt this analysis below. Whereas our proposed shifted sampler and post-training strategy improve both PSNR and rFID simultaneously, the naive sampling strategy described above only allows us to trade off rFID and PSNR against each other, constrained by the ``rate-distortion-perception tradeoff'' \cite{ratedistortionperception}. Importantly, rFID and PSNR cannot both be arbitrarily increased. See Figure \ref{fig:ratedistortionperception_a} for a quantitative study of this process. Where the $x_1$ likelihood is increased too high, the result is an eventual \textit{degradation} in the sample's perceptual quality, while PSNR continues to increase. 
Even if it were possible to maximize the sample likelihood, as noted in \cite{universalcritic}, directly maximizing sample likelihood is inappropriate for image realism.

\subsection{Dataset and resolution generalization.} FlowMo-Lo is a diffusion autoencoder, so it naturally generalizes to higher resolutions by employing a patchwise diffuse-and-blend strategy known in prior work \cite{highfidelity}. \modelname-Lo achieves strong performance on unseen data, at 256 and 512 resolution, in terms of rFID, despite a lower BPP and only training on ImageNet-1K at $256\times 256$. In Table \ref{tab:multires}, we compare with the Cosmos DI-8x8 tokenizer, a generalizable tokenizer trained at multiple resolutions on a large internet-scale dataset.

\begin{table}[h]
\vspace{-2.5mm}
    \footnotesize
    \centering
    \begin{tabular}{l|ccccccc}
        \midrule
        & \multicolumn{2}{c}{ImageNet-1K} 
        & \multicolumn{2}{c}{OpenImages}
        & \multicolumn{2}{c}{Food-101} \\
        \midrule
        Resolution & 256 & 512 & 256 & 512 & 256 & 512\\
        \midrule
        Cosmos DI-8x8 & 0.87 & 0.39 & 0.85 & 0.52 & \textbf{1.05} & 0.94 \\
        FlowMo-Hi & \textbf{0.55} & \textbf{0.30} & \textbf{0.69} & \textbf{0.40} & 1.39 & \textbf{0.68} \\
        \midrule
    \end{tabular}
    \caption{\label{tab:multires}\textbf{Multiple resolutions.} FlowMo can perform encoding and decoding at arbitrary resolutions following a patchwise diffusion strategy known in prior work \cite{highfidelity}.}
    
\end{table}

\subsection{Tokenizer scaling.} \modelname~is a transformer-only architecture parameterized in $\mu P$ to facilitate easy scaling. We show in the table below that, holding all else constant, \textit{all metrics improve every time the width factor is increased}. For this experiment we train for 200K steps only. Other config settings are the same as FlowMo (fewer params).

\begin{table}[h]
    \footnotesize
    \centering
    \begin{tabular}{cc|ccc}
         \midrule
         $\mu P$ width & \# Params ($\times 10^6)$ & rFID $\downarrow$ & PSNR $\uparrow$ & 
         LPIPS $\downarrow$ \\
         \midrule
         2 & 260 & 7.77 & 20.84 & 0.169 \\
         3 & 367 & 5.31 & 21.28 & 0.160 \\
         4 & 517 & 4.45 & 21.60 & 0.155 \\
         5 & 710 & \textbf{3.84} & \textbf{21.70} & \textbf{0.152} \\
         \midrule
    \end{tabular}
    \label{tab:scaling}
\vspace{-4mm}
\end{table}

\subsection{``Mode-seeking,'' precision and recall} 
We illustrate the `mode-seeking' effect of post-training below. We compare the statistics of decompressed images to reference dataset statistics via the Precision and Recall metrics \cite{precrec}. When there is a 1:1 correspondence between the elements of the conditioning dataset and the reference dataset, post-training improves both precision and recall. This corresponds to reducing the diversity of the distribution $p(x|c)$ for each $c$ by concentrating around its modes, which improves fidelity to the original reference dataset.

By contrast, when comparing to the train set statistics (i.e. an unrelated set of images not in 1:1 correspondence with the conditioning dataset), post-training improves precision (quality) at the cost of recall (diversity).

\begin{table}[h]
    \footnotesize
    \centering
    \begin{tabular}{l|cc|cc}
    \midrule
         & \multicolumn{2}{c|}{vs. train stats.} & \multicolumn{2}{c}{vs. val stats.} \\
         Model name & Prec.$\uparrow$ & Rec.$\uparrow$ & Prec.$\uparrow$ & Rec.$\uparrow$ \\
         \midrule
         FlowMo (fewer params) (no posttrain)& 0.734 & \textbf{0.660} & 0.974 & 0.988 \\
         FlowMo (fewer params) & \textbf{0.766} & 0.634 & \textbf{0.993} & \textbf{0.991} \\
         \midrule
    \end{tabular}
    \label{tab:scaling}
\end{table}

\subsection{More results with post-training} The post-training scheme proposed generalizes to multiple architectures, consistently improving their performance. 
We have added more results extending Table 5 below, for FlowMo (fewer params), which is defined in Table \ref{tab:training_hyperparameters}, and FlowMo-Continuous, which is the version against which we compare DiTo in Table 3.

\begin{table}[h]
\vspace{-3mm}
    \footnotesize
    \centering
    \begin{tabular}{l|ccc}
    \midrule
         Model name & rFID $\downarrow$ & PSNR $\uparrow$ & LPIPS $\downarrow$ \\
         \midrule
         FlowMo (fewer params) (no post-train)& 2.02 & 21.32 & 0.143 \\
         FlowMo (fewer params) & \textbf{1.21} & \textbf{22.20} & \textbf{0.120} \\
         \midrule
         FlowMo-Continuous (no post-train) & 0.74 & 25.65 & 0.064 \\
         FlowMo-Continuous  & \textbf{0.65} & \textbf{26.61} & \textbf{0.054} \\
         \midrule
    \end{tabular}
\vspace{-4.5mm}
\end{table}

\subsection{Wall clock times} We provide wall-clock times below. 
Note that FlowMo encoding is comparable to LlamaGen-32, so it does not bottleneck second-stage generative modeling. The main slowdown is due to a diffusion decoder, which we consider acceptable because it enables significantly better reconstruction. Decoding could be made faster ($\approx$\hspace{-0.025mm} $5\times$ to $10\times$) via distillation.
\begin{table}[h]
    \footnotesize
    \centering
    \begin{tabular}{l|ccc}
    \midrule
         Model name & Encode 1 image (s) & Decode 1 image (s) \\
         \midrule
         LlamaGen-32 & 0.021 & \textbf{0.038} \\
         FlowMo-B & 0.050 & 2.391 \\
         FlowMo-B (small) & \textbf{0.011} & 0.322 \\
     \midrule
    \end{tabular}
    \label{tab:scaling}
\end{table}

\section{Extended visualization}
We provide extended comparisons between our method and various state-of-the-art tokenizers in Figures \ref{fig:comparison1a}, \ref{fig:comparison1b}, \ref{fig:comparison2a}, \ref{fig:comparison2b} below. Images are not cherry-picked.

\clearpage

\begin{figure*}
    \centering
    \begin{tabular}{c@{}c@{}c@{}}
    \includegraphics[width=.29\linewidth]{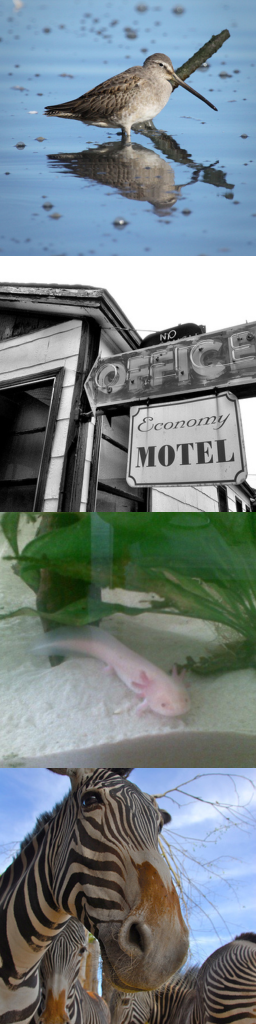} & 
    \includegraphics[width=.29\linewidth]{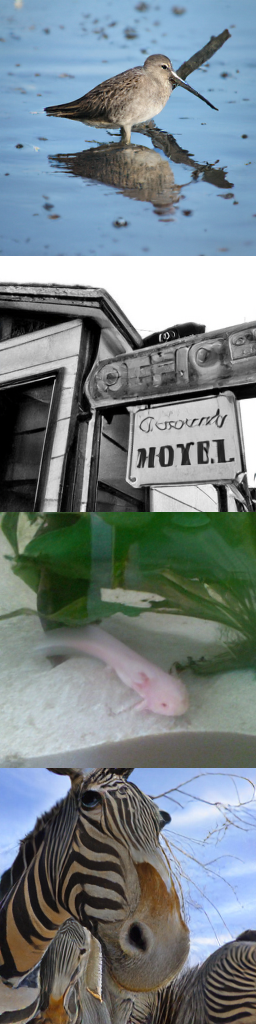} & 
    \includegraphics[width=.29\linewidth]{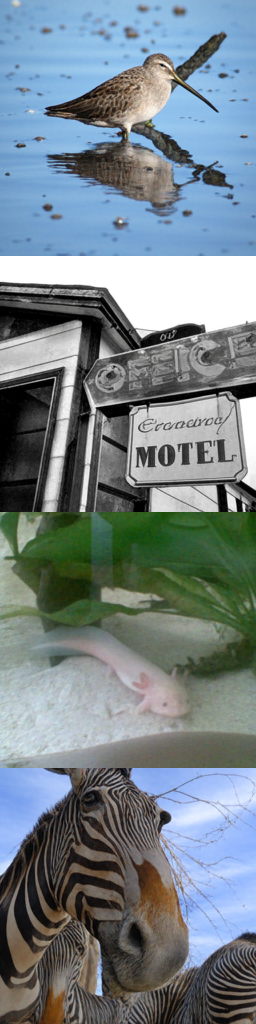} \\
    Original image & Reconstructed & Reconstructed \\
    & (OpenMagViT-V2 \cite{openmagvit_v2}) & (\modelname-A)
    \end{tabular}
    \caption{\label{fig:comparison1a} Tokenizer comparison.}
\end{figure*}

\begin{figure*}
    \centering
    \begin{tabular}{c@{}c@{}c@{}}
    \includegraphics[width=.29\linewidth]{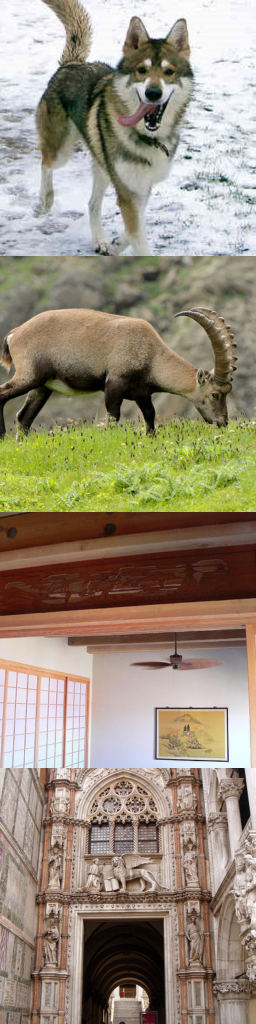} & 
    \includegraphics[width=.29\linewidth]{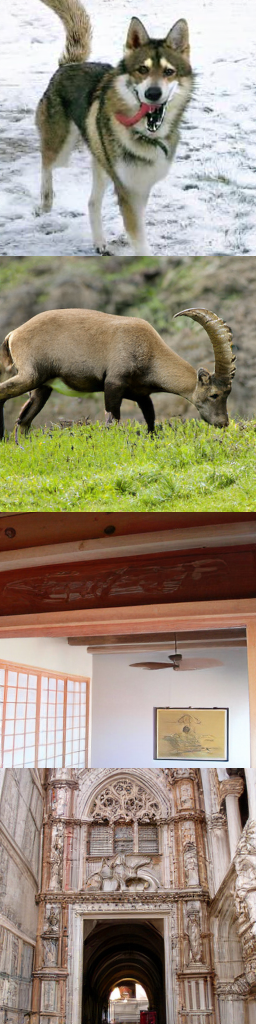} & 
    \includegraphics[width=.29\linewidth]{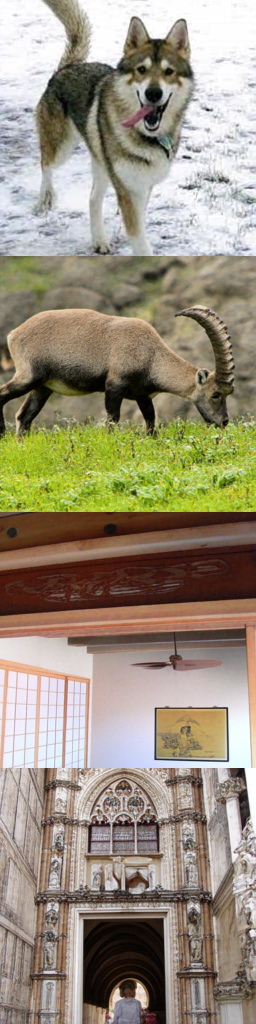} \\
    Original image & Reconstructed & Reconstructed \\
    & (OpenMagViT-V2 \cite{openmagvit_v2}) & (\modelname-A)
    \end{tabular}
    \caption{\label{fig:comparison1b} Tokenizer comparison, continued.}
\end{figure*}

\begin{figure*}
    \centering
    \begin{tabular}{c@{}c@{}c@{}}
    \includegraphics[width=.29\linewidth]{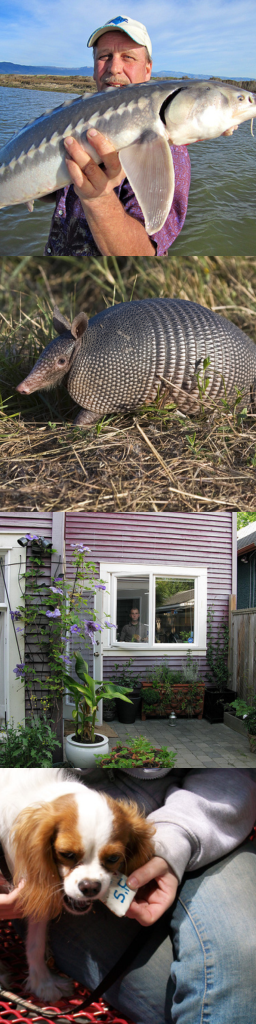} & 
    \includegraphics[width=.29\linewidth]{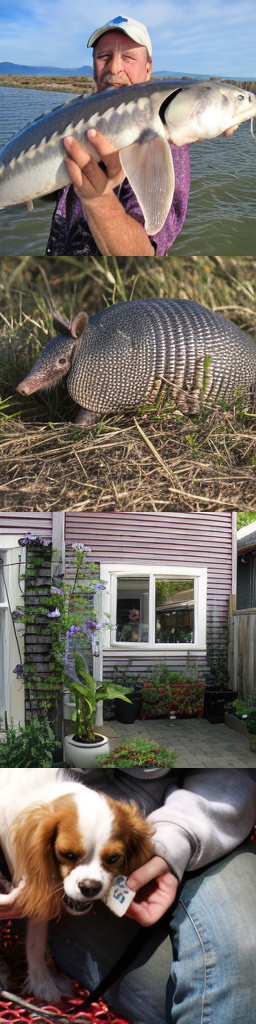} & 
    \includegraphics[width=.29\linewidth]{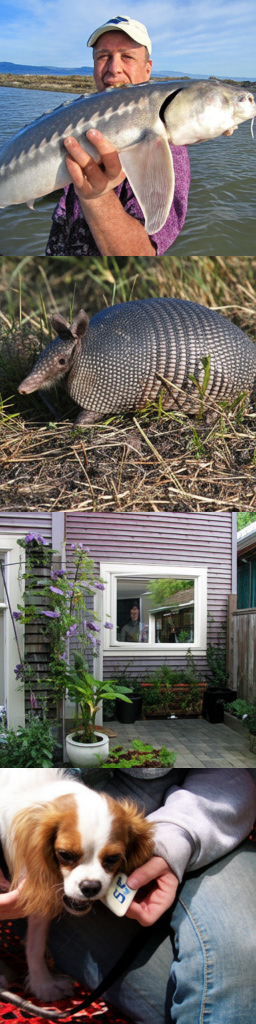} \\
    Original image & Reconstructed & Reconstructed \\
    & (LlamaGen-32 \cite{llamagen}) & (\modelname-B)
    \end{tabular}
    \caption{\label{fig:comparison2a} Tokenizer comparison, continued.}
\end{figure*}

\begin{figure*}
    \centering
    \begin{tabular}{c@{}c@{}c@{}}
    \includegraphics[width=.29\linewidth]{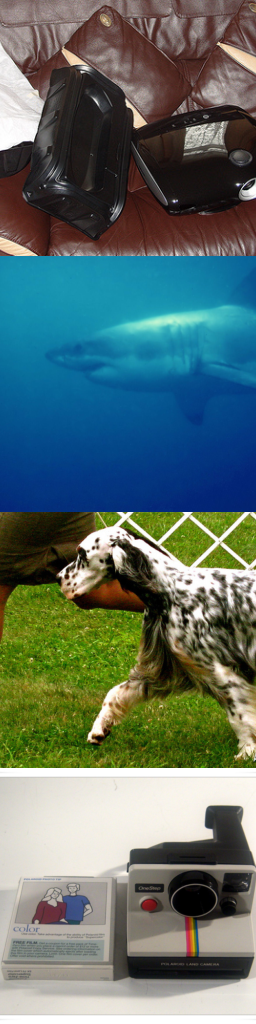} & 
    \includegraphics[width=.29\linewidth]{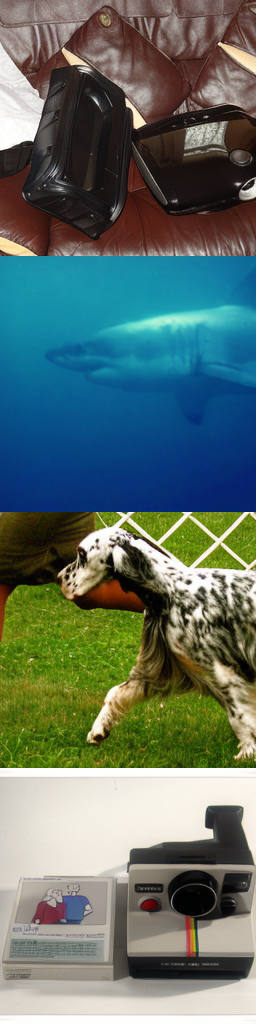} & 
    \includegraphics[width=.29\linewidth]{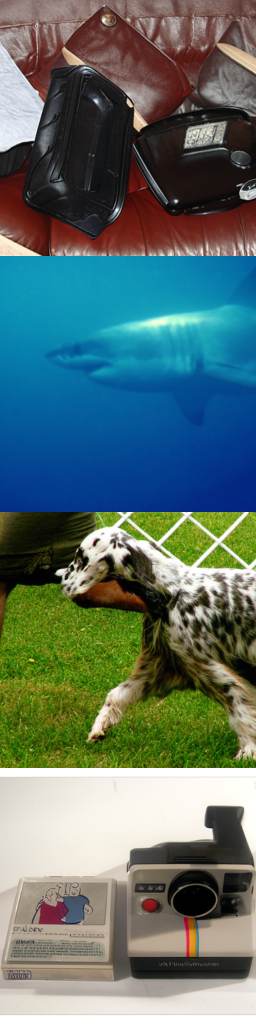} \\
    Original image & Reconstructed & Reconstructed \\
    & (LlamaGen-32 \cite{llamagen}) & (\modelname-B)
    \end{tabular}
    \caption{\label{fig:comparison2b} Tokenizer comparison, continued.}
\end{figure*}

\clearpage

\end{document}